\documentclass[lettersize,journal]{IEEEtran}
\usepackage{amsmath,amsfonts}
\usepackage{algorithmic}
\usepackage{algorithm}
\usepackage{array}
\usepackage[caption=false,font=normalsize,labelfont=sf,textfont=sf]{subfig}
\usepackage{textcomp}
\usepackage{stfloats}
\usepackage{url}
\usepackage{verbatim}
\usepackage{graphicx}
\usepackage{cite}
\usepackage{multirow}
\usepackage{graphicx}
\usepackage{amsmath}
\usepackage{amsthm}
\usepackage{booktabs}
\usepackage{subfiles}
\usepackage{color}
\usepackage{pdfpages}
\usepackage{float}
\usepackage{wrapfig}
\usepackage{picinpar}
\usepackage{cutwin}
\usepackage{comment}
\usepackage{float}  
\usepackage{lipsum}
\usepackage{amssymb}
\usepackage[multiple]{footmisc}

\begin{document}

\title{ModelLight: Model-Based Meta-Reinforcement Learning for Traffic Signal Control}

\author{Xingshuai Huang\IEEEmembership{,}
        Di Wu~\IEEEmembership{Member,~IEEE,} \\
        Michael Jenkin~\IEEEmembership{Senior Member,~IEEE}
        and~Benoit Boulet~\IEEEmembership{Senior Member,~IEEE} 
\thanks{Xingshuai Huang, Di Wu, Benoit Boulet are with the Department of Electrical and Computer Engineering, McGill University; Email: xingshuai.huang@mail.mcgill.ca, di.wu5@mcgill.ca, benoit.boulet@mcgill.ca.\\ Di Wu is the corresponding author.}
\thanks{Michael Jenkin is with the department of Electrical Engineering and Computer Science, York University (e-mail: jenkin@eecs.yorku.ca).}

}



\maketitle

\begin{abstract}

Traffic signal control is of critical importance for the effective use of transportation infrastructures. The rapid increase of vehicle traffic and changes in traffic patterns make traffic signal control more and more challenging. Reinforcement Learning (RL)-based algorithms have demonstrated their potential in dealing with traffic signal control. However, most existing solutions require a large amount of training data, which is unacceptable for many real-world scenarios. This paper proposes a novel model-based meta-reinforcement learning framework (ModelLight) for traffic signal control. Within ModelLight, an ensemble of models for road intersections and the optimization-based meta-learning method are used to improve the data efficiency of an RL-based traffic light control method. Experiments on real-world datasets demonstrate that ModelLight can outperform state-of-the-art traffic light control algorithms while substantially reducing the number of required interactions with the real-world environment.

\end{abstract}


\section{Introduction}

Traffic congestion has become an important issue in many cities around the world.  Efficient traffic signal control has been treated as one of the most promising and feasible solutions to alleviate traffic congestion and can significantly reduce the average travel time, hence increasing city mobility and promoting urban economic growth~\cite{wei2018intellilight}. Given the potential advantages of effective signal control, traffic signal control has been a hot research area for many years~\cite{Qadri:2020}.  Early approaches based on statistical modelling of historical traffic flow have been replaced with more responsive models based on measuring cars queuing for and passing through an intersection. 

Reinforcement learning (RL)-based approaches have shown a number of advantages when compared with traditional traffic signal control methods which use pre-defined fixed control schemes~\cite{gartner1995development,lowrie1990scats}. Studies typically focus on applying RL algorithms to different traffic situations, such as a single intersection \cite{DBLP:conf/nips/OroojlooyNHS20,zheng2019learning,zang2020metalight}, roundabout \cite{rizzo2019time}, and multi-intersection \cite{chen2020toward,yumacar,wang2020large} conditions. Most RL-based methods are based on model-free reinforcement learning (MFRL) algorithms, which rely on direct interaction with the real environment to learn the control policy. The sample efficiency of such approaches is low and thus a large amount of training data and a large number of training iterations are required to learn an efficient control policy. Given the complicated traffic condition of a real intersection, massive real training data is infeasible, and excessive training time is also intolerable \cite{zang2020metalight}. Thus, it is important to improve data efficiency for learning-based traffic control algorithms. 

In order to overcome the requirement of a large number of training episodes, MetaLight \cite{zang2020metalight}, a meta-reinforcement learning-based method, was proposed.  MetaLight improves the learning efficiency of the task by transferring knowledge learned previously from other tasks to the target task. However, MetaLight still requires a large number of interactions with the real environment in other tasks, which is time-consuming and not necessarily possible for real-world applications. 

Model-based reinforcement learning (MBRL), a method utilizing one or more learned dynamics models of the environment to assist the policy learning, has been shown to be more sample efficient~\cite{clavera2018model}. Many previous studies in MBRL focus on solving model bias \cite{clavera2018model,chua2018deep,kurutach2018model} or developing theoretical bounds for MBRL while only a few studies apply MBRL to practical issues \cite{lambert2019low,nagabandi2018learning,kaiser2019model,zhang2019solar}, and the existing applications are typically robotics-specific. This paper utilizes the MBRL method for the traffic signal control problem. Specifically, we learn an ensemble of dynamic models of the environment using neural networks to generate imaginary transitions so as to mitigate the reliance on interactions with the real environment. MetaLight~\cite{zang2020metalight}, a meta-reinforcement learning framework based on MAML \cite{finn2017model} is also adopted in our method to learn an optimal initialization of the parameters which can be tuned to target tasks. Furthermore, FRAP++ networks \cite{zang2020metalight} are adopted to approximate the Q function during policy training. 

The contributions of this paper can be summarized as follows:
\textbf{(i)} To the best of our knowledge, this is the first attempt that applies MBRL to the traffic signal control problem.
Specifically, we propose a novel model-based meta-reinforcement learning (MBMRL) method named ModelLight for traffic signal control.
\textbf{(ii)} The feasibility and advantages of our proposed method are demonstrated by rigorous experiments on real-world datasets. Results show that our method can obtain state-of-the-art traffic signal control performance while significantly reducing the required number of interactions with the real environment.
\section{Related Work}
\subsection{Reinforcement Learning-Based Traffic Signal Control}
There are two main types of traffic signal control methods: pre-timed and adaptive methods \cite{genders2020policy}. As an example of a pre-timed approach, Webster’s method \cite{webster1958traffic} is a representative controller which generates fixed phase setting and cycle duration without considering the current traffic state. 

With the growth and reduced cost of computing power,  adaptive traffic signal control has become the norm. Given its suitability for RL-based control, RL-based control approaches are one of the most active research directions in the adaptive traffic signal control field. Simple RL algorithms, such as Sarsa \cite{thorpe1996tra} and Q-learning \cite{abdulhai2003reinforcement}, were among the RL algorithms applied to traffic signal control. Given increasing numbers of traffic states, deep RL, which uses deep neural networks to approximate the value or policy function, shows its potential in solving more complex traffic control problems. For value-based deep RL, neural networks are utilized to map states to the value function so as to output a deterministic policy. For instance, Deep Q Network (DQN), a typical value-based deep RL algorithm, is widely studied in traffic signal control \cite{zheng2019learning}.
Policy-based deep RL exploits neural networks to learn a policy that directly outputs actions or distribution of actions based on the states \cite{rizzo2019time}. Actor-critic algorithms, which take advantage of both value function and policy gradient-based RL, have also attracted considerable attention in the traffic signal control field \cite{yang2019cooperative}.

Many works 
leverage state-of-the-art machine learning techniques to further improve the performance of an RL-based controller. \cite{yumacar} utilizes Message Propagation-based Graph Neural Networks (MPGNN) to form a Communication Agent Network (CAN) which outputs coordinated actions in a multi-intersection traffic signal control problem. Colight \cite{wei2019colight} employs graph attention networks to learn weights for adjacent intersections thus assisting communication between different RL agents. PlanLight~\cite{zhang2020planlight} proposes to use behaviour cloning to address the concerns on reward design for traffic light control. Meta-learning is also exploited in RL-based traffic signal control. MetaLight \cite{zang2020metalight} adopts enhanced MAML ~\cite{finn2017model} to learn a general initial policy based on some existing tasks, hence accelerating the training process of an RL-based traffic signal controller. \cite{zhu2021meta} applies meta-reinforcement learning to traffic signal control for multi-intersections. To our knowledge, however,  existing RL methods applied to traffic signal control are all MFRL-based, which shows low sample efficiency and requires a considerable amount of training data to learn a good policy. \cite{zhang2020generalight} proposed a Wasserstein generative adversarial network-based traffic flow generator to improve the model generalization of traffic light control agent over different traffic flow scenarios.  

\subsection{Model-Based Reinforcement Learning (MBRL)}
With prior knowledge of how an environment works, MBRL addresses MFRL's requirement that considerably larger amounts of training data are required than are required for human learning \cite{kaiser2019model}. Recent publications on MBRL include both theoretical and practical work. For the most well-known theoretical MBRL frameworks, \cite{langlois2019benchmarking} proposed three categories: Dyna-style algorithms, shooting algorithms, and policy search with backpropagation through time. Model-Based Meta-Policy-Optimization (MB-MPO) \cite{clavera2018model}, Stochastic Lower Bound Optimization (SLBO) \cite{luo2018algorithmic} and Model-Ensemble Trust-Region Policy Optimization (ME-TRPO) \cite{kurutach2018model} are typical Dyna-style algorithms which are built based on Dyna frameworks \cite{sutton1990integrated}. These approaches use imaginary transitions generated from the learned dynamics model to replace data sampled from the real environment so as to reduce the interaction with the environment. Shooting algorithms adopt the idea of Model Predictive Control (MPC) to handle non-convex reward optimization and nonlinear systems. Examples include the Random Shooting (RS) \cite{rao2009survey}, Probabilistic Ensembles with Trajectory Sampling (PETS) \cite{chua2018deep}, and Mode-Free Model-Based (MB-MF) \cite{nagabandi2018neural} algorithms. Probabilistic Inference for Learning Control (PILCO) \cite{deisenroth2011pilco}, Guided Policy Search (GPS) \cite{chebotar2017path}, and Iterative Linear Quadratic-Gaussian (iLQG) \cite{tassa2012synthesis} approaches belong to the last category which exploit both data and gradient analysis to guide value or policy optimization.

In addition to theoretical studies, MBRL has also shown superior performance in practical applications. For instance, the SOLAR method \cite{zhang2019solar} shows better performance than other MBRL approaches and is more efficient than MFRL on robotic tasks. \cite{lambert2019low} utilized MBRL to realize low-level control of the hover of a quadrotor with less training data, accelerating the speed of production of low-level controllers. MBRL has also been employed in complex tasks such as Atari games. SimPLe \cite{kaiser2019model} outperformed MFRL methods with a lower data regime. Recently, \cite{huang2019learning} employed MBRL with a neural renderer to learn to paint like a human painter.  Experimental results showed its feasibility on various target images.

In our work, the MBRL algorithm and meta-learning method are combined to improve the performance of traffic signal control and enhance sample efficiency, mitigating the training burden in the real environment.

\section{Preliminaries}
\subsection{Traffic Signal Control Problem}
\begin{figure}[t]
    \centering
  \subfloat[\tiny][Standard intersection with 8 incoming lanes.\label{fig1a}]{%
      \includegraphics[width=0.45\linewidth]{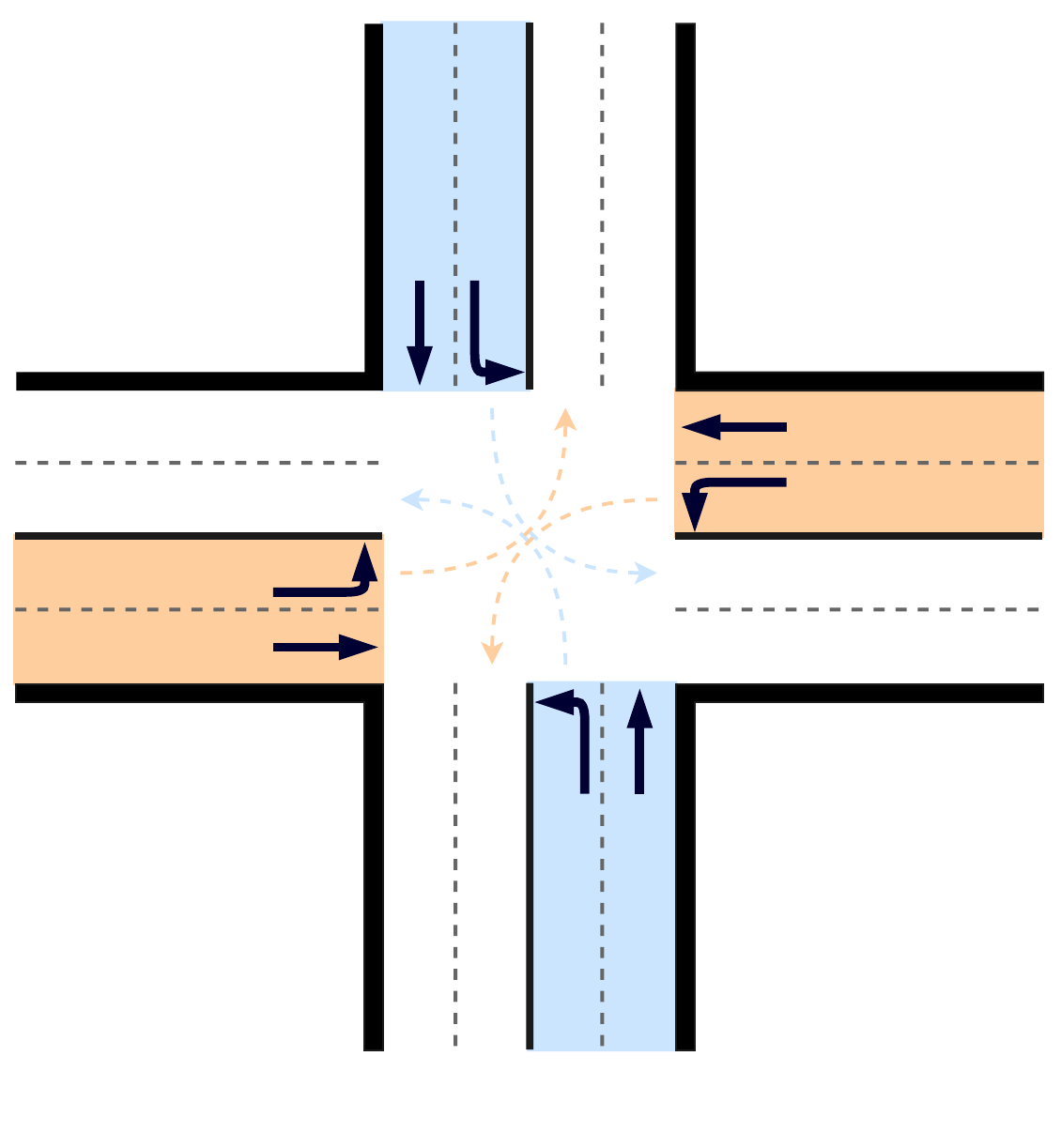}}
    \hfill
  \subfloat[8 primary phases.\label{fig1b}]{%
        \includegraphics[width=0.45\linewidth]{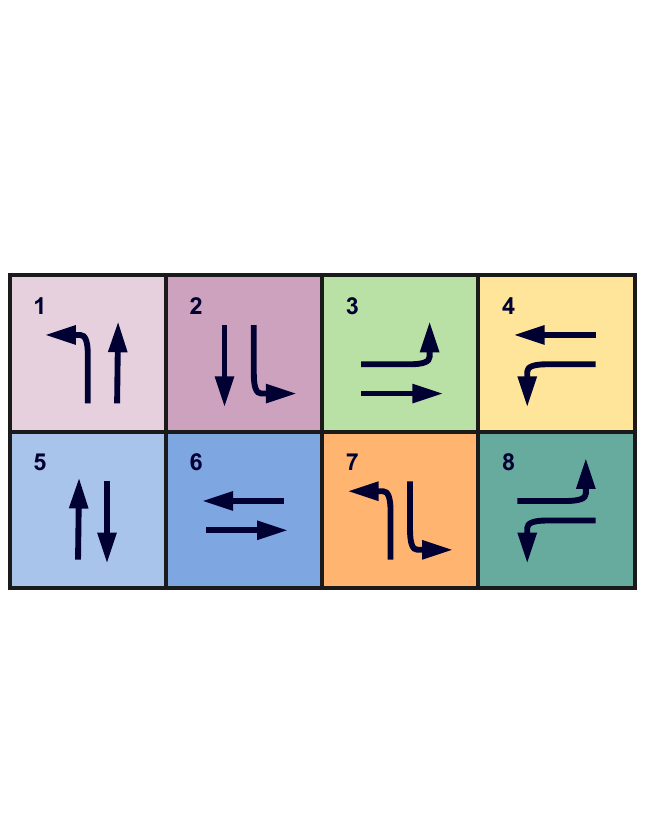}}
  \caption{Standard intersection with 8 incoming lanes and primary phases. (a) shows the structure of the intersection while (b) shows the primary traffic phases of traffic flow. Right-hand turn lanes are typically not modelled in traffic flow models.}
  \label{fig:fig1} 
\end{figure}

This paper considers the problem of traffic signal control of a  standard single intersection consisting of four two-lane approaches, as illustrated in Figure~\ref{fig1a}. The model includes different approaches, lanes, traffic flows, and phases.

\textbf{Approaches and Lanes} 
An intersection is composed of multiple approaches with one or more lanes in each approach. Different types of lanes represent different vehicle movements. For instance, as illustrated in Figure~\ref{fig1a}, two different lanes restrict the vehicles on respective lanes to turn left or go straight. 

\textbf{Traffic Flow} Traffic flow refers to the number of vehicles passing through the cross-section of a road section in unit time. 

\textbf{Phases} A phase is an association of different traffic signals in different lanes at the same time, which controls the orderly movement of vehicles at the intersection and prevents conflicts. As summarized in Figure~\ref{fig1b}, there are eight different phases in this particular intersection. Each phase permits the movement of two lanes while restricting the remaining lanes.

\subsection{Model-Based Reinforcement Learning for Traffic Signal Control}
The process of traffic signal control is modeled as a Markov Decision Process (MDP) $\langle\mathcal{S}, \mathcal{A}, \mathcal{P}, \mathcal{R}, \gamma\rangle$, where $\mathcal{S}$ and $\mathcal{A}$ represent the state space and action space. $\mathcal{P}$ is the state transition and $\mathcal{R}$ denotes the reward function. $\gamma$ is the discount factor. Unlike MFRL algorithms whose state transition $\mathcal{P}$ and reward function $\mathcal{R}$ are unknown, MBRL methods first learn a model of the environment $Model(S, A)$ and then leverage the learned model to improve value or policy optimization, hence boosting sample efficiency. In the traffic signal control problem, the reward function $\mathcal{R}$ and the state transition $\mathcal{P}$ are learned using supervised learning with some real transitions $\{(s,a,s',r)\}$, where $s,a,s',r$ are the current state, selected action, next state, and reward, respectively.

\section{Methodology}
This paper employs model-based reinforcement learning (MBRL)  to learn an ensemble of models of the intersection environment and then uses meta learning to learn the optimal initialization parameters that fit target tasks well with a few training steps. In this section, the MDP formulation is first introduced. Next, we present the learning process of the ensemble of intersection models, followed by the model-based meta-reinforcement learning (MBMRL) framework adopted in this study. Finally, the implementation details are described.

\subsection{Markov Decision Process Formulation}
The traffic light control problem can be formulated as a Markov Decision Process. Here, we present the detailed definitions for the state, action, and reward for our problem. 

\textbf{State} In a real intersection, the environment state contains a series of variables. Following the MetaLight approach~\cite{zang2020metalight}, this study considers the two most important state variables: (i) The number of vehicles waiting on each incoming lane except the right-turn lane since vehicles in the right-turn lane can generally turn right without being restricted by traffic light signals. (ii) The current phase of the traffic signal. 

\textbf{Action} The only variables in an intersection model that are actively controlled are the alternation and duration of traffic lights on each incoming approach. For simplicity, this paper uses the phase selection for the next time period as the agent action. For a standard intersection with 4 two-lane approaches, the action space contains 8 actions (phases). 

\textbf{Reward} Restricted by the delay in obtaining travel time, in this case, we take the opposite of the queue length $q_{k}$ on each incoming lane -- an equivalent substitution of travel time~\cite{zheng2019diagnosing}, as the reward variable:
$
 {R}=-q_{k}
$.

\subsection{Learning of Intersection Models}
For MBRL algorithms, the choice of the type of model which represents the dynamics of the environment is pivotal since the corresponding control policy will be influenced significantly by any flaw in the model~\cite{abbeel2006using}. Taking into account the characteristics of traffic flow, we choose long short-term memory (LSTM) to learn each model, since this architecture has shown good performance on time series prediction problems.
 
We use a function $M_{\phi}$ with parameters $\phi$ to represent the intersection model which predicts the reward $r$ and the next state $s'$ according to the current state $s$ and selected action $a$. The learning objective for each model $M_{\phi_{i}}$ is to find the optimal $\phi_{i}$ that minimize the following $L_{2}$ loss:
 \begin{equation}
 \label{equ2}
     \min _{\phi_{i}} \frac{1}{\left|\mathcal{D}_{i}\right|} \sum_{\left(\boldsymbol{s}, \boldsymbol{a}, \boldsymbol{s'}, \boldsymbol{r}\right) \in \mathcal{D}_{i}}\left\|(\boldsymbol{s'}, \boldsymbol{r})-M_{\boldsymbol{\phi}_{i}}\left(\boldsymbol{s}, \boldsymbol{a}\right)\right\|^{2}
 \end{equation}
 \noindent
where $\mathcal{D}_{i}$ is the subset of randomly sampled transitions from the transition set $\mathcal{B}_{i}$.
To further reduce model bias caused by deficiencies of the learned environment model, we adopt an ensemble of dynamic models of the environment~\cite{clavera2018model}; We randomly initialize the parameters of every model and train different models with different samples of the transitions to reduce the correlation between different models. Additionally, new transitions sampled from the real environment are continuously used to retrain every model hence preventing the distributional shift problem~\cite{clavera2018model}.

\subsection{Meta-Reinforcement Learning on Learned Intersection Models}
Meta-learning, also known as learning to learn, aims to enable the machine learning agent to learn fast. Model-Agnostic Meta-Learning (MAML) is a well-known meta-learning method that which aims to learn a good model initialization \cite{finn2017model}. In this work, we propose to tackle the traffic signal control problem with a MAML-based meta-learning MetaLight (Zang et al. 2020), and model-based RL. 
With the ensemble of intersection models $\left\{M_{\phi_{1}},M_{\phi_{2}},...,M_{\phi_{N}}\right\}$, we formulate each traffic signal control task as an MDP which is expressed as $\langle\mathcal{S}, \mathcal{A}, M_{\phi_{i}}, \gamma\rangle$. 

We utilize a function $f_{\theta}$ with parameters $\theta$ to represent the meta policy (meta learner) which maps the current state to the action output. Accordingly, for each task, the adapted policy (base learner) is defined as the function $f_{\theta_{i}^{\prime}}$ with parameters $\theta_{i}^{\prime}$. The meta-training objective is to learn the initialization parameters $\theta$ to optimize the performance of each adapted policy $f_{\theta_{i}^{\prime}}$ on the corresponding task, which can be expressed as minimizing the following loss:

\begin{algorithm}[H]
  \caption{Meta Training of ModelLight}
  \label{a1}
\begin{algorithmic}
  \small
  \STATE {\bfseries Require:} Distribution over training tasks $p(\mathcal{T})$; adaptation step size $\alpha$; meta step size $\beta$; meta update frequency $m$; number of real transitions $T$; number of imaginary transitions $N$; number of meta training rounds $H$ \\
  \STATE Randomly initialize meta policy $f_{\theta}$ with parameter $\theta$ 
  \FOR {$h=1,2,...,H$}
  \STATE Randomly sample batch of tasks $\mathcal{T}_{i} \sim p(\mathcal{T})$; initialize transition set $\mathcal{B}_{i}$ to $\emptyset$
  \FOR {$t=1,m+1,...,T$}
            \STATE $\theta'_{i} \leftarrow \theta$
            \FOR {$t'=t,t+1,...,\min{(t+m-1,T)}$}
                \FOR {Each task $\mathcal{T}_{i}$}
                    \STATE Generate real transitions into $\mathcal{B}_{i}$ and sample transitions $\mathcal{D}_{i}$ using $f_{\theta'_{i}}$
                    \STATE Evaluate $\nabla_{\theta'_{i}} \mathcal{L}_{\mathcal{T}_{i}}(f_{\theta'_{i}})$ using transitions $\mathcal{D}_{i}$ and Equation~\ref{equ1}
                    \STATE Compute updated parameters with gradient descent $\theta_{i}^{\prime} \leftarrow \theta'_{i}-\alpha \nabla_{\theta'_{i}} \mathcal{L}_{\mathcal{T}_{i}}\left(f_{\theta'_{i}}\right)$ 
                \ENDFOR
            \ENDFOR
            \STATE Sample transitions $\mathcal{D'}_{i}$ from $\mathcal{B}_{i}$
            \STATE Update $\theta \leftarrow \theta-\beta \nabla_{\theta} \sum_{i} \mathcal{L}_{\mathcal{T}_{i}}\left(f_{\theta_{i}^{\prime}}\right)$ using transitions $\mathcal{D}_{i}^{\prime}$ and Equation~\ref{equ3} and ~\ref{equ1}
        \ENDFOR
        \STATE Learn intersection models $M_{\phi_{i}}$ with randomly sampled subset $\mathcal{D}_{i}''$ from $\mathcal{B}_{i}$ to minimize $\min _{\phi_{i}} \frac{1}{\left|\mathcal{D''}_{i}\right|} \sum_{\left(\boldsymbol{s}, \boldsymbol{a}, \boldsymbol{s'}, \boldsymbol{r}\right) \in \mathcal{D''}_{i}}\left\|(\boldsymbol{s'}, \boldsymbol{r})-M_{\boldsymbol{\phi}_{i}}\left(\boldsymbol{s}, \boldsymbol{a}\right)\right\|^{2}$
        \FOR {$t=1,m+1,...,N$}
            \STATE $\theta'_{i} \leftarrow \theta$
            \FOR {$t'=t,t+1,...,\min{(t+m-1,N)}$}
                \FOR {Each intersection model $M_{\phi_{i}}$}
                    \STATE Generate imaginary transitions into $\mathcal{B}_{i}$ and sample transitions $\mathcal{E}_{i}$ using $f_{\theta'_{i}}$
                    \STATE Evaluate $\nabla_{\theta'_{i}} \mathcal{L}_{M_{\phi_{i}}}(f_{\theta'_{i}})$ using transitions $\mathcal{E}_{i}$ and Equation~\ref{equ1}
                    \STATE Compute adapted parameters with gradient descent $\theta_{i}^{\prime} \leftarrow \theta'_{i}-\alpha \nabla_{\theta'_{i}} \mathcal{L}_{M_{\phi_{i}}}\left(f_{\theta'_{i}}\right)$
                \ENDFOR
            \ENDFOR
            \STATE Sample imaginary transitions $\mathcal{E}_{i}^{\prime}$ from $\mathcal{B}_{i}$
            \STATE Update $\theta \leftarrow \theta-\beta \nabla_{\theta} \sum_{i} \mathcal{L}_{M_{\phi_{i}}}\left(f_{\theta_{i}^{\prime}}\right)$ using transitions $\mathcal{E}_{i}^{\prime}$ and Equations~\ref{equ3} and ~\ref{equ1}
        \ENDFOR
  \ENDFOR
\end{algorithmic}
\end{algorithm}

\begin{equation}
\label{equ3}
    \min _{\theta} \mathcal{L}(f_{\theta}) =\sum_{i=1}^{N} \mathcal{L}_{M_{\phi_{i}}}\left(f_{\theta_{i}^{\prime}}\right) 
\end{equation}
\noindent

With $\mathcal{L}_{M_{\phi_{i}}}\left(f_{\theta_{i}^{\prime}}\right)$ being the loss under the adapted policy $\theta_{i}^{\prime}$ and the intersection model $M_{\phi_{i}}$.

\begin{equation}
\begin{split}
\label{equ1}
    \begin{array}{l}
        \mathcal{L}_{M_{\phi_{i}}}\left(f_{\theta_{i}^{\prime}}\right)= 
        \mathbb{E}_{a \sim f_{\theta_{i}^{\prime}}(a|s);s',r \sim M_{\phi_{i}}(s,a)}\\ \left[\left(r+\gamma \max \limits_{a^{\prime}} Q\left(s^{\prime}, a^{\prime} ; f_{\hat{\theta}_{i}}\right)-Q\left(s, a ; f_{\theta_{i}^{\prime}}\right)\right)^{2}\right] 
    \end{array}
\end{split}
\end{equation}

\noindent
where $Q(s^{\prime}, a^{\prime} ; f_{\hat{\theta}_{i}})$ denotes the value function of the next action $a'$ under the next state $s'$ for a given policy $f_{\hat{\theta}_{i}}$, and $\hat{\theta}_{i}$ is the parameters of the target network. $Q\left(s, a ; f_{\theta_{i}^{\prime}}\right)$ is the value function of the selected action $a$ under the current state $s$ for policy $f_{\theta_{i}^{\prime}}$.

We use imaginary transitions generated by each intersection model $M_{\phi_{i}}$ with adapted policy $f_{\theta'_{i}}$ to update the parameters $\theta_{i}^{\prime}$ of adapted policy, which can be formulated as 
$
    \theta_{i}^{\prime} \leftarrow \theta'_{i}-\alpha \nabla_{\theta'_{i}} \mathcal{L}_{M_{\phi_{i}}}\left(f_{\theta'_{i}}\right)
$, where
$\alpha$ is the adaptation step size. 
We then update the initialization parameters $\theta$ of meta policy with newly sampled imaginary transitions:
\begin{equation}
    \theta \leftarrow \theta-\beta \nabla_{\theta} \sum_{i=1} \mathcal{L}_{M_{\phi_{i}}}\left(f_{\theta_{i}^{\prime}}\right)
\end{equation}
\noindent where $\beta$ denotes the meta step size.

\subsection{Algorithm Implementation} 

In this paper, we propose ModelLight to tackle the traffic signal control problem. There are two main phases for ModelLight, i.e., meta training and meta testing phases. Pseudo-code for the meta training phase of ModelLight is presented in Algorithm \ref{a1}. The meta training phase can be further split into 3 parts: (i) meta training in real world, (ii) learning intersection models, and (iii) meta training in intersection models. We first randomly select a batch of training tasks $\mathcal{T}_{i}$ from the training task distribution $p(\mathcal{T})$. (i) Then, we employ the meta policy $f_{\theta}$ with initialization parameters in real world $\theta$ to obtain real transitions into a transition set $\mathcal{B}_{i}$ for each task $\mathcal{T}_{i}$. The adapted parameters $\theta'_{i}$ are then updated using some sampled real transitions $\mathcal{D}_{i}$. When updating the initialization parameters $\theta$, some new transitions $\mathcal{D'}_{i}$ are sampled from the transition set $\mathcal{B}_{i}$ and used to calculate the gradients. (ii) After interacting with the real environment, we learn an ensemble of intersection models ${M_{\phi_{i}}}$ using some newly sampled real transitions $\mathcal{D''}_{i}$. (iii) In the subsequent process, the learned intersection models are utilized to generate imaginary transitions for another meta training iteration which are similar to the meta training in real world. In the meta testing phase, we apply the meta policy with the learned optimal initialization parameters $\theta$ for adaptation training. 
With a small number of interactions with the real environment and gradient updates, the desired adapted policy with the updated adapted parameters will be obtained.

In the implementation, we use a value-based algorithm, i.e., Deep Q Network (DQN) \cite{mnih2015human}, to optimize our policy. FRAP++ \cite{zang2020metalight}, an improved FRAP model which is composed of embedding and convolutional layers, is used as the value network. The approximators we adopt to represent intersection models are LSTMs with a shared network structure, where one head outputs predicted $s'$, and another generates $r$. The network consists of an input layer with 16 input nodes, a two-layer LSTM with 16 and 64 hidden nodes, and a one-layer fully-connected neural network with 9 output nodes. The Adam optimizer is used to train the networks. The number of models in the ensemble is the same as the number of total different training tasks, and we use real transitions from each task to train the corresponding model. The adaptation and meta step size are both set as 0.001. The numbers of real transitions and imaginary transitions for each episode are both 360, which represents 3600s interaction with the environment. In meta test phase, the initial model will be adaptively trained for 1 episode. To ensure reproducibility, more hyperparameter details are given in Table~\ref{h1}. Inspired by \cite{kaiser2019model}, we apply the short rollouts technique to reduce compounding errors in learned models. Every 36 time steps we randomly sample a starting state from the transition set when generating imaginary transitions via the learned intersection models. Differing from some Dyna-style MBRL algorithms, in this study, the sampled real transitions are also exploited for meta training to further mitigate compounding errors.

\noindent
\begin{table}
\centering
\caption{Values of hyperparameters for reproduction.}
\label{h1}
\begin{tabular}{p{6.6cm}p{0.6cm}}
\toprule
\textbf{Hyperparameter} & \textbf{Value} \\
\hline
Number of meta training rounds $H$                            & 100    \\
Number of training epochs for LSTM                       & 200    \\
Number of randomly sampled tasks per round               & 2      \\
Number of real transitions per task per round ($T$)        & 360    \\
Number of imaginary transitions per task per round ($N$)   & 360    \\
Number of sampled real transitions for model training   & 300    \\
Batch size of real transitions for policy training      & 30     \\
Batch size of imaginary transitions for policy training & 100    \\
Adaptation step size $\alpha$                                   & 0.001  \\
Meta step size $\beta$                                           & 0.001  \\
Meta update frequency $m$                                  & 10     \\
Optimizer for intersection model training~               & Adam   \\
Optimizer for Q-network training~                        & SGD    \\
Initial epsilon parameter of $\epsilon$-greedy                    & 0.8    \\
Minimum epsilon parameter of $\epsilon$-greedy                    & 0.2    \\
Updating frequency of target network                     & 5  \\ 
\bottomrule
\end{tabular}
\end{table}

\section{Experiments}

Experiments were conducted using the commonly used simulator, CityFlow\footnote{https://cityflow-project.github.io}, on a cloud cluster with 2 x Intel E5-2683 v4 Broadwell @ 2.1GHz CPU. 
In this work, we use CityFlow to simulate the real environment. The transitions generated from simulator are referred to as real transitions, which represent interactions with the real environment. The transitions generated by the learned intersection models are referred to as imaginary transitions \cite{clavera2018model}.
These experiments aim to answer several questions: \textbf{Q1} Does the ModelLight boost learning efficiency and lead to better performance on tasks similar to training tasks? \textbf{Q2} Does ModelLight also improve learning efficiency and performance on tasks distinct from the training tasks? \textbf{Q3} Does ModelLight well addresses tasks with varying action spaces? \textbf{Q4} How does ModelLight compare to state-of-the-art meta-learning methods? \textbf{Q5} Does ModelLight further mitigate reliance on interactions with the real environment?

\begin{figure}[htb]
    \centering
    \vspace{-5pt}
    \includegraphics[width=8cm]{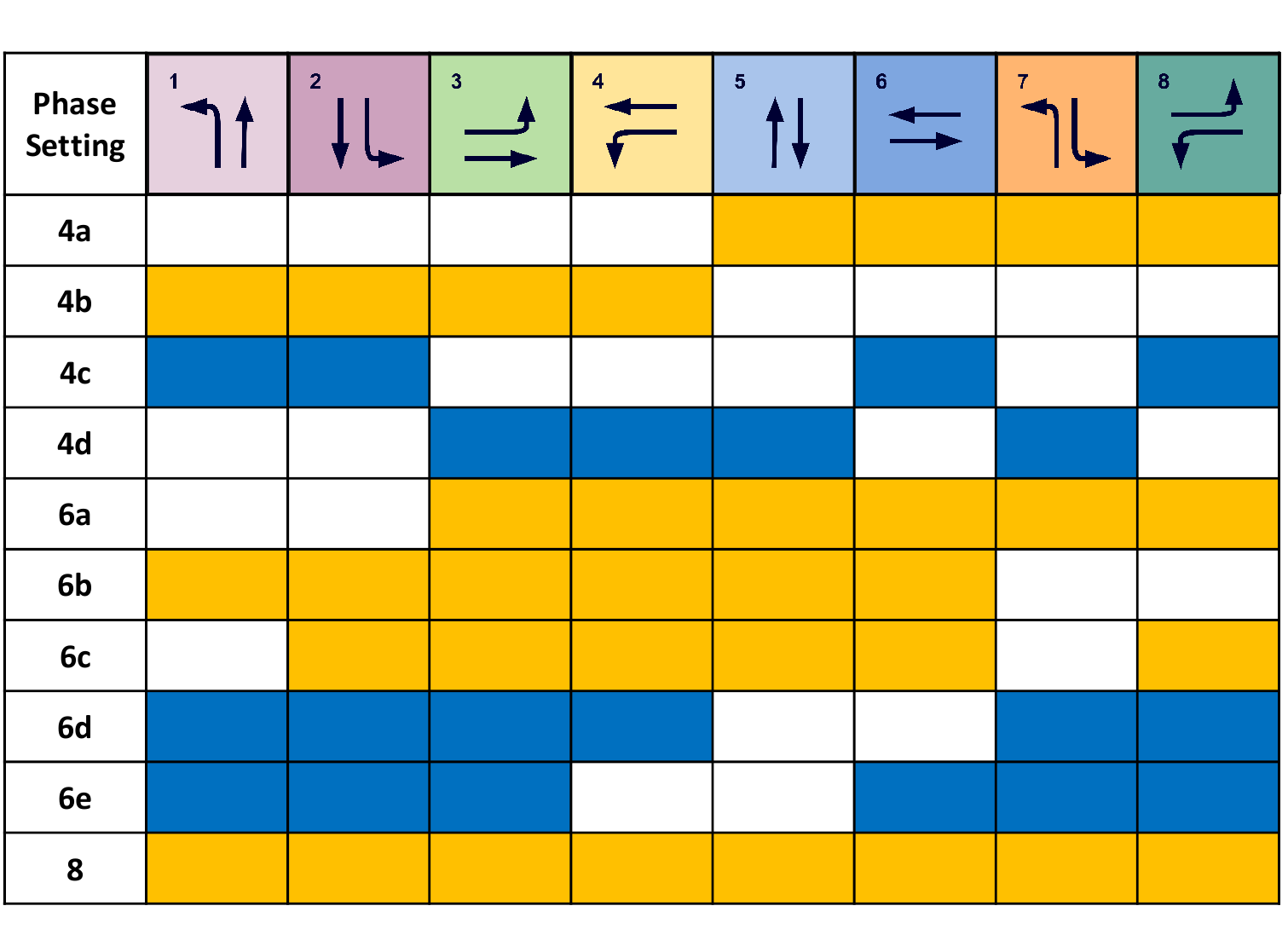}
    \vspace{-5pt}
    \caption{Ten different phase settings. Yellow represents phase setting 1 (PS1) and blue represents phase setting 2 (PS2).}
    \vspace{-8pt}
    \label{fig3}
\end{figure}

Considering the comprehensive and sophisticated experimental settings of MetaLight \cite{zang2020metalight}, we employ four real-world datasets (traffic flow and roadnet data) from it; $D_{\text{Jinan}}, D_{\text{Hangzhou}},D_{\text{Atlanta}}$, and $D_{\text{LosAngeles}}$, which are from four different cities: Jinan, Hangzhou, Atlanta, and LosAngeles, respectively. 
We also adopt a number of the phase settings from MetaLight. Specifically, as shown in Figure~\ref{fig3}, intersection with ten different phase settings were used in the experiment, including one 8-phase intersection, four different 4-phase intersections, and five different 6-phase intersections. In order to handle the varying size of the action space caused by different phase settings and solve potential intersections with different lanes, we follow the settings of FRAP++ \cite{zang2020metalight} by sharing parameters of embedding layers across lanes and fixing the number of $1 \times 1$ convolutions.

In the meta training phase, 20 different intersection scenarios with phase setting (PS) 1 from $D_{\text{Hangzhou}}$ are utilized to train our policy, including 3$*$4a, 3$*$4b, 3$*$6a, 3$*$6b, 3$*$6c and 5$*$8, where 4a, 4b, 6a, 6b, 6c and 8 are phase settings of PS1. To evaluate how ModelLight performs, we use 3 task settings in meta test phase. In Task 1, 12 intersection scenarios with PS1 from $D_{\text{Hangzhou}}$ are adopted, which studies the performance on scenarios similar to meta training scenarios except for traffic flow. Task 2 is conducted for the evaluation of performance on scenarios from different cities but with the same phase settings as meta training scenarios, where six different intersection scenarios with PS1 from $D_{\text{Jinan}}, D_{\text{Atlanta}}$, and $D_{\text{LosAngeles}}$ are used for meta testing. In Task 3, we further test how ModelLight performs on  a challenging setting. Scenarios are sampled from different cities with different phase settings, where 10 intersection scenarios with PS2 from $D_{\text{Jinan}}, D_{\text{Atlanta}}$, and $D_{\text{LosAngeles}}$ are used for meta testing.

To verify the superiority and effectiveness of ModelLight on traffic signal control problems, several state-of-the-art baselines are used as comparison.
\begin{itemize}
    \item \textbf{FRAP++} \cite{zang2020metalight} is the adapted policy of ModelLight. FRAP++ is directly applied to the 3 test tasks with random initialization parameters.
    \item\textbf{MAML} \cite{finn2017model} combines the original FRAP with MAML framework.
    \item \textbf{MetaLight} \cite{zang2020metalight} learns a good initialization of FRAP++ using an adapted MAML method and is shown to be superior than other baselines in most of the traffic scenarios.
\end{itemize}

The travel time, a widely adopted traffic measurement which is defined as the average travel time that vehicles spend on approaching lanes (in seconds), is used as the evaluation index to evaluate the performance of different methods under different experimental scenarios.

\subsection{Adapted Performance}
In this experiment, we evaluate how ModelLight compares to other traffic signal control methods when using the same number of interactions with simulator (real transitions) on 3 tasks. 100 training rounds (100 pass over all the dataset) are applied for meta-learning methods (MAML, MetaLight, and ModelLight) and a 3600-second test is conducted for all methods.

Table \ref{T1} and \ref{T2} reports the performance (average travel time $\pm$ standard deviation) of different methods on 3 tasks. Lower average travel time with lower standard deviation denotes better performance. MetaLight is the second-best algorithm and shows better performance when compared with FRAP++ and MAML in most cases. ModelLight outperforms all baselines in five out of six cases of Task 1 and all six cases of Task 2 and 3, yielding average improvements of 5.71\%, 17.94\% and 9.69\% in terms of average travel time for Task 1, 2 and 3 respectively when compared with the best baseline (\textbf{Q3, Q4}). Furthermore, ModelLight achieves the lowest standard deviation in the vast majority of the scenarios, which implies the superior reliability of ModelLight. In most cases, given that model error can significantly influence the final performance of model-based RL, MFRL methods generally achieve similar or better asymptotic performance as MBRL as long as the number of training rounds is not limited \cite{chua2018deep, nagabandi2018learning}. However, our MBRL method can outperform MFRL method when the number of training round is limited considering that an infinite amount of training is not feasible in a real intersection. Additionally, MAML generates similar results as FRAP++, while the other two meta-learning methods, MetaLight and ModelLight, greatly surpass the performance of FRAP++ on both tasks, which addresses \textbf{Q1} and \textbf{Q2}.

\begin{figure} 
    \centering
  \subfloat[Phase setting: 4a\label{fig3a}]{%
       \includegraphics[width=0.45\linewidth]{ 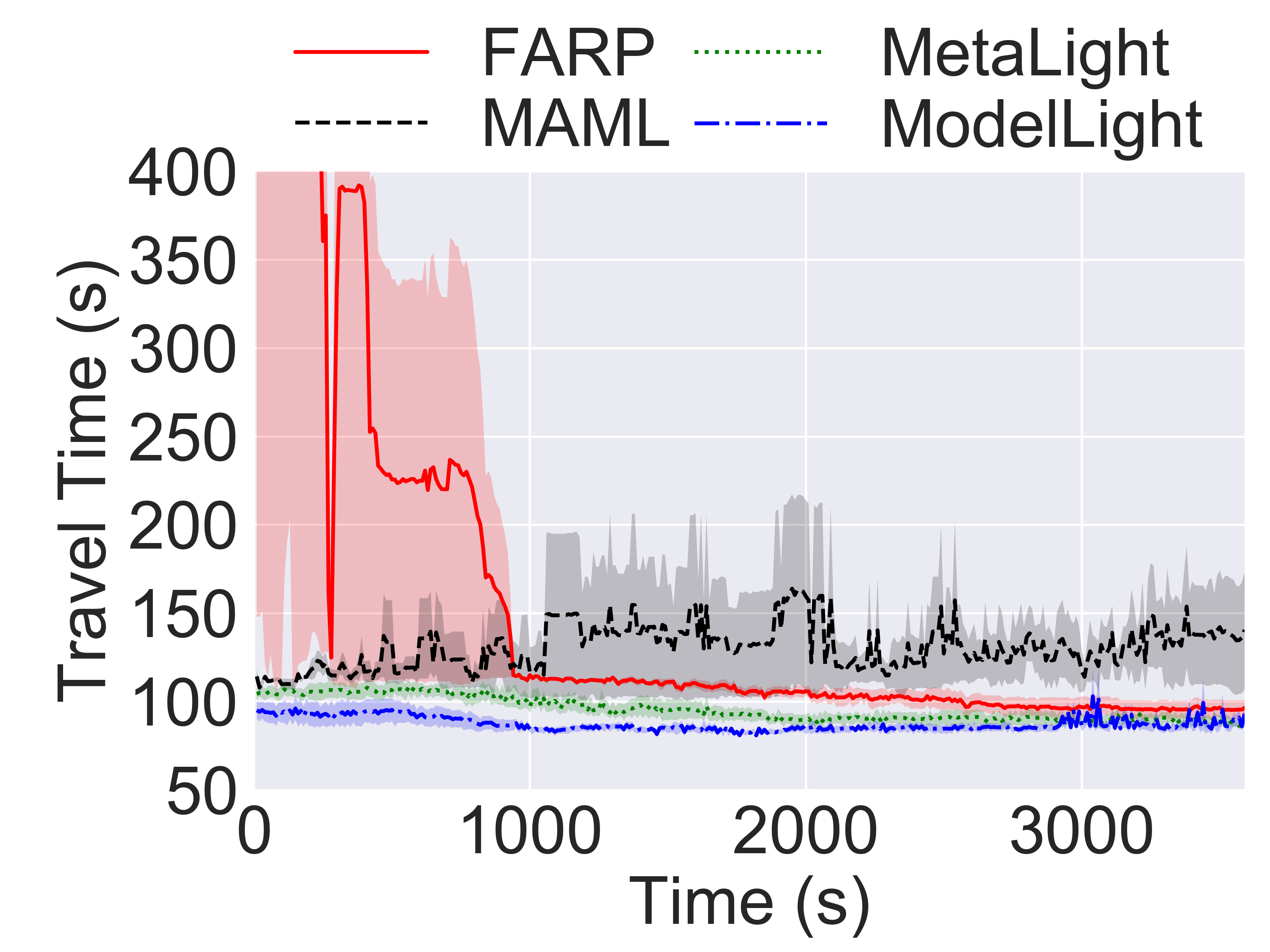}}
    \hfill
  \subfloat[Phase setting: 4b\label{fig3b}]{%
        \includegraphics[width=0.45\linewidth]{ 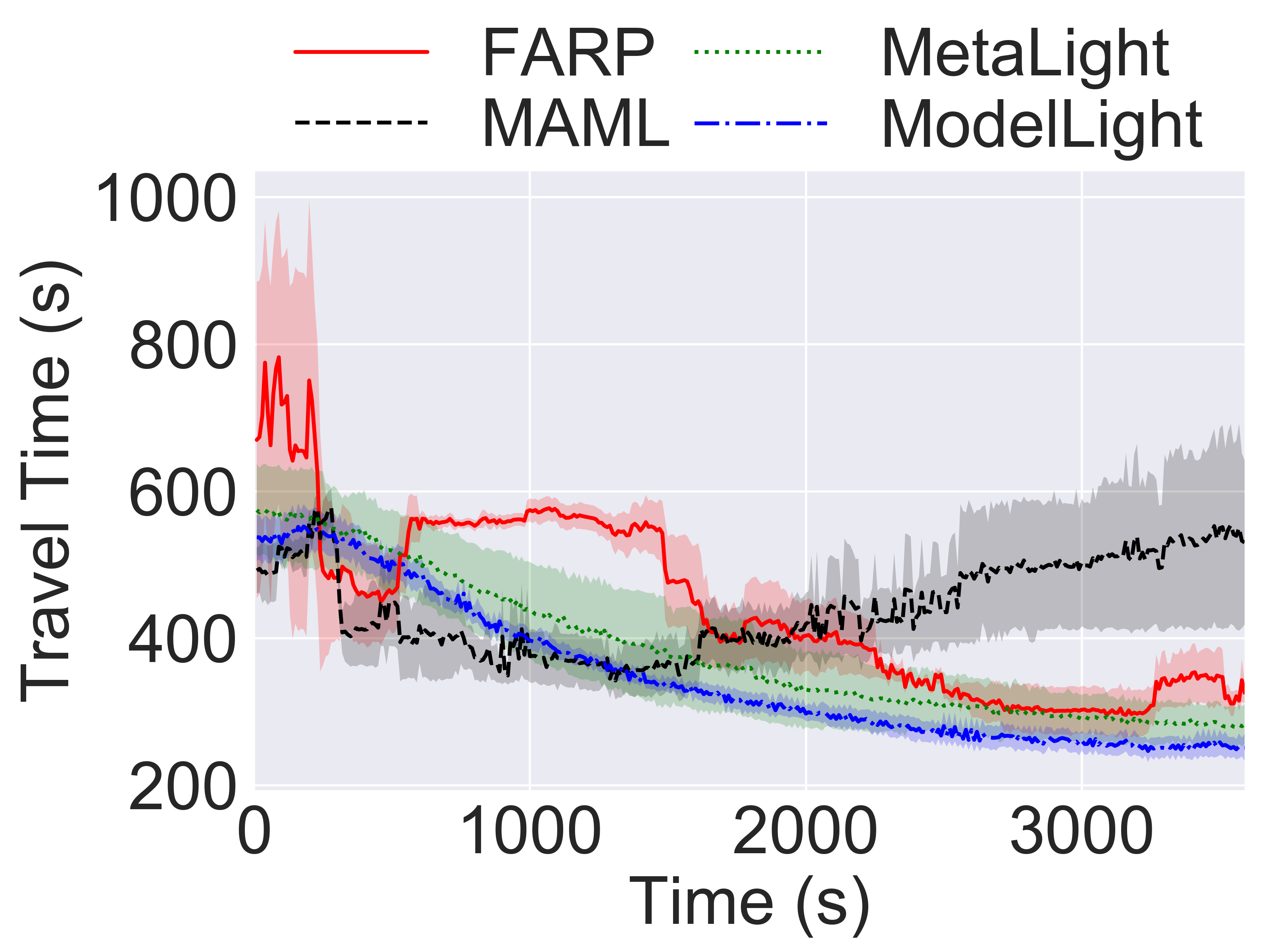}}
    \\
  \subfloat[Phase setting: 6a\label{fig3c}]{%
        \includegraphics[width=0.45\linewidth]{ 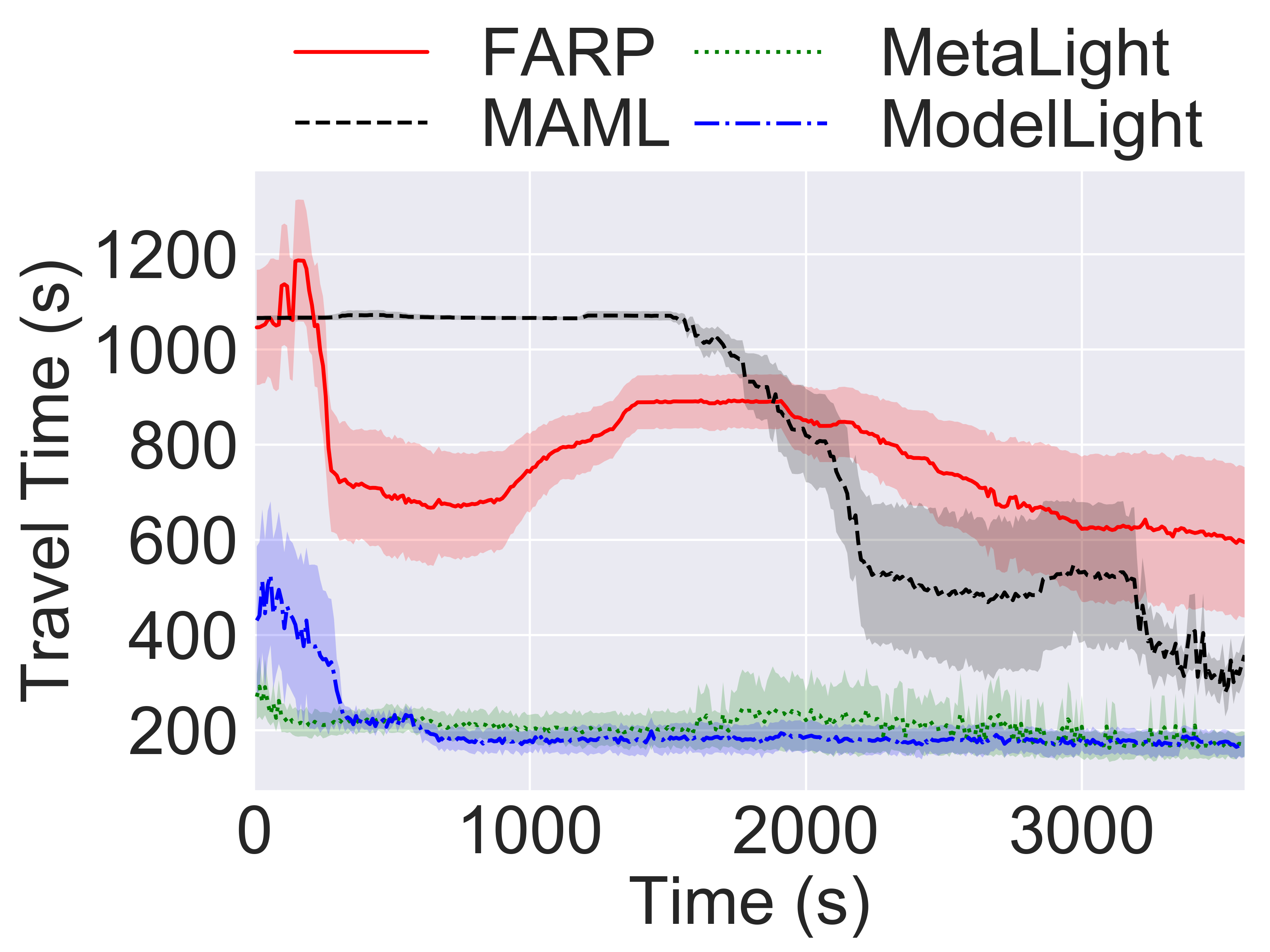}}
    \hfill
  \subfloat[Phase setting: 6b\label{fig3d}]{%
        \includegraphics[width=0.45\linewidth]{ 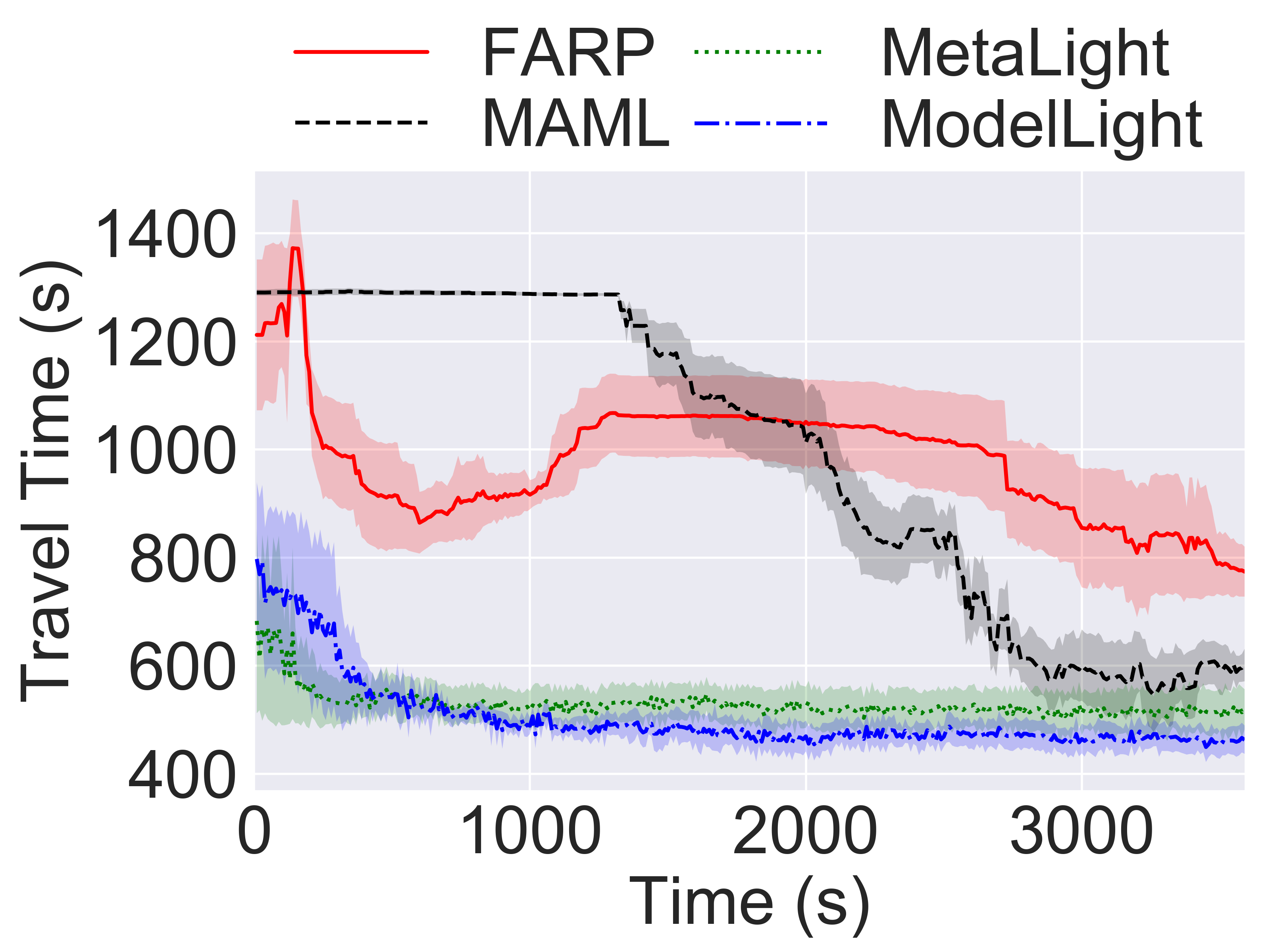}}
  \caption{Comparison of adapted performance on Task 1. The duration of an episode is 3600s. Travel time curves with means and variances of 5 random seeds are illustrated.}
  \label{fig 3} 
\end{figure}

\begin{figure} 
    \centering
  \subfloat[Scenario: LA 4a\label{fig4a}]{%
       \includegraphics[width=0.45\linewidth]{ 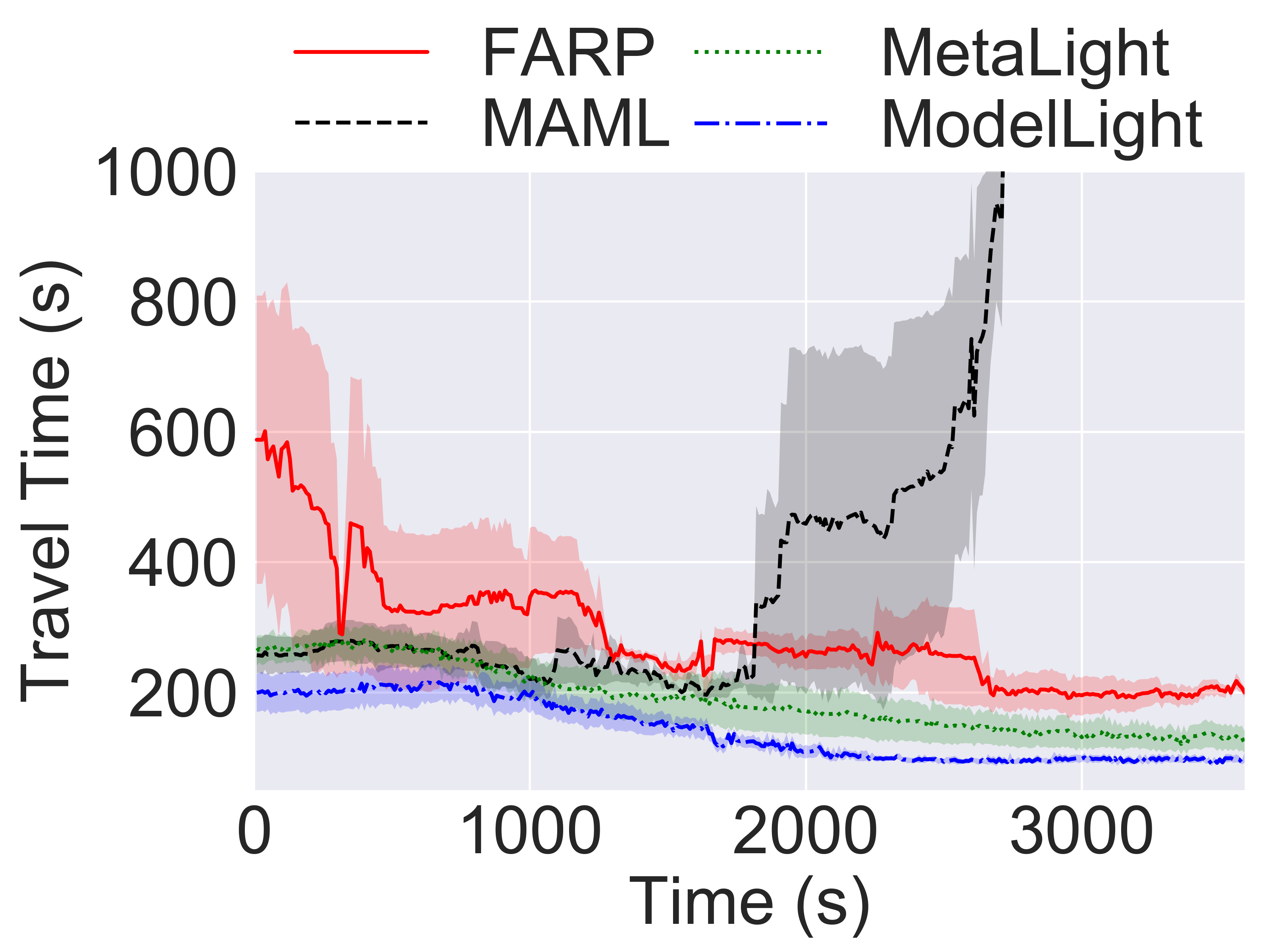}}
    \hfill
  \subfloat[Scenario: LA 6d\label{fig4b}]{%
        \includegraphics[width=0.45\linewidth]{ 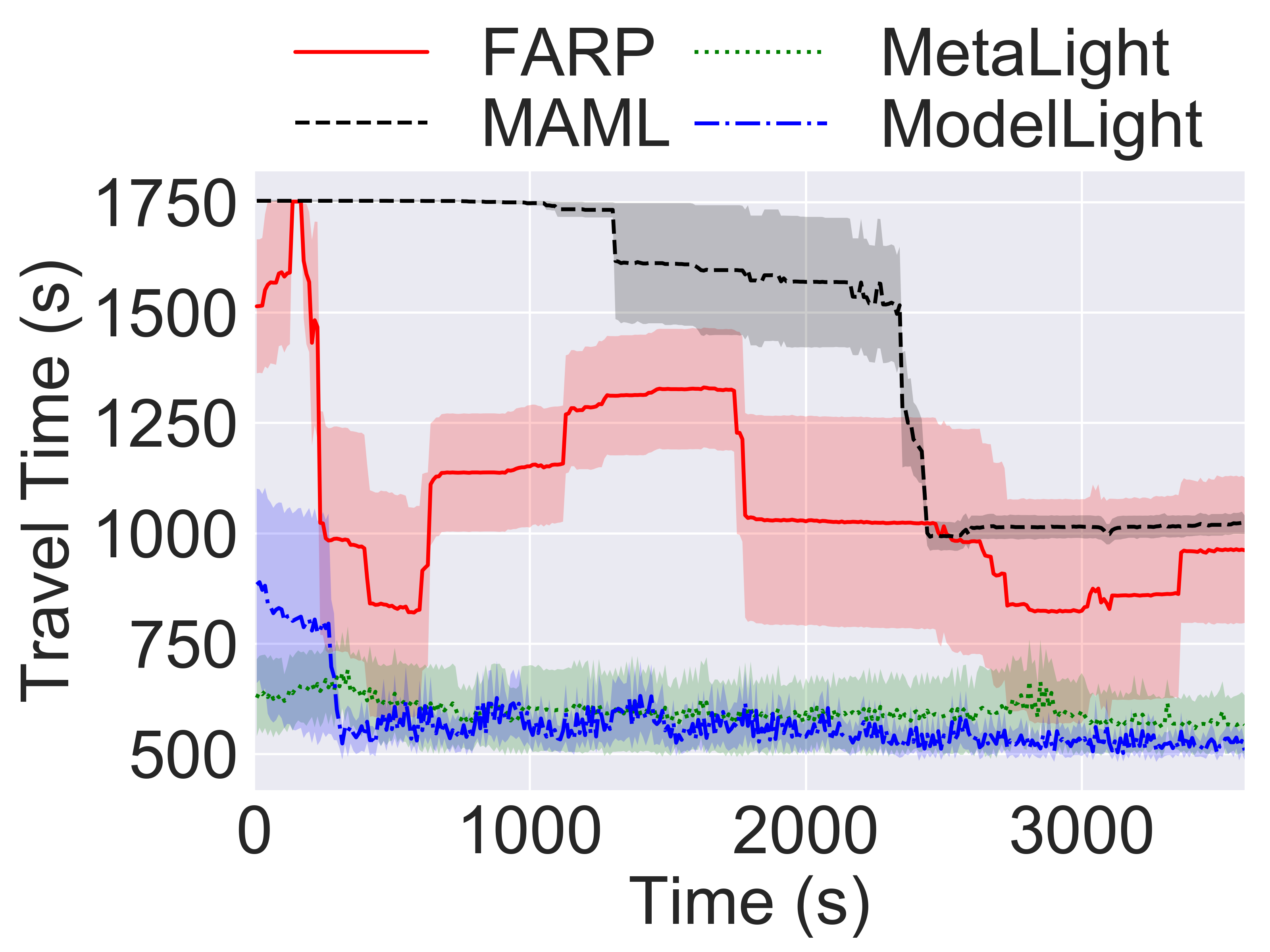}}
    \\
  \subfloat[Scenario: AT 4d\label{fig4c}]{%
        \includegraphics[width=0.45\linewidth]{ 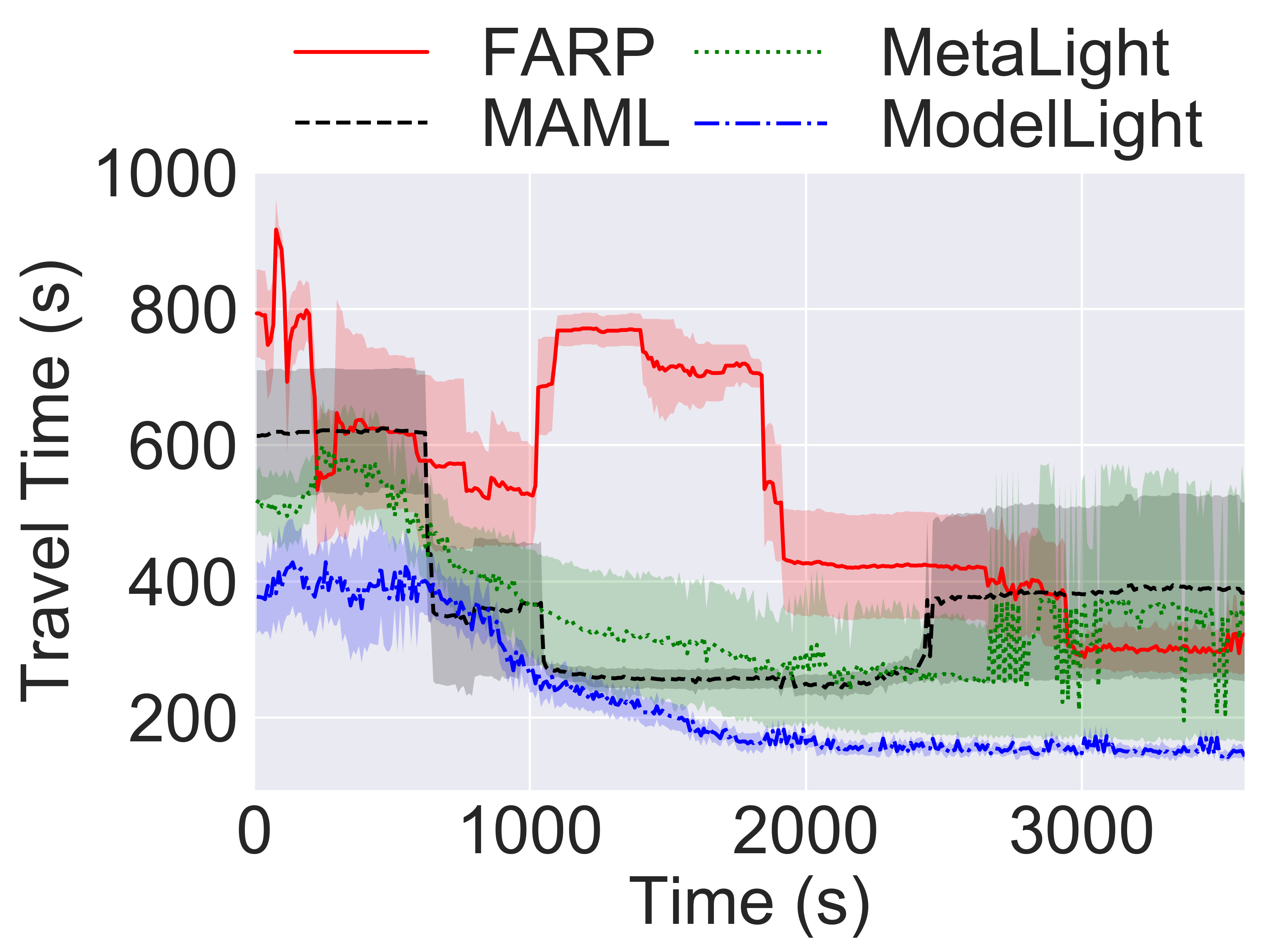}}
    \hfill
  \subfloat[Scenario: JN 4c\label{fig4d}]{%
        \includegraphics[width=0.45\linewidth]{ 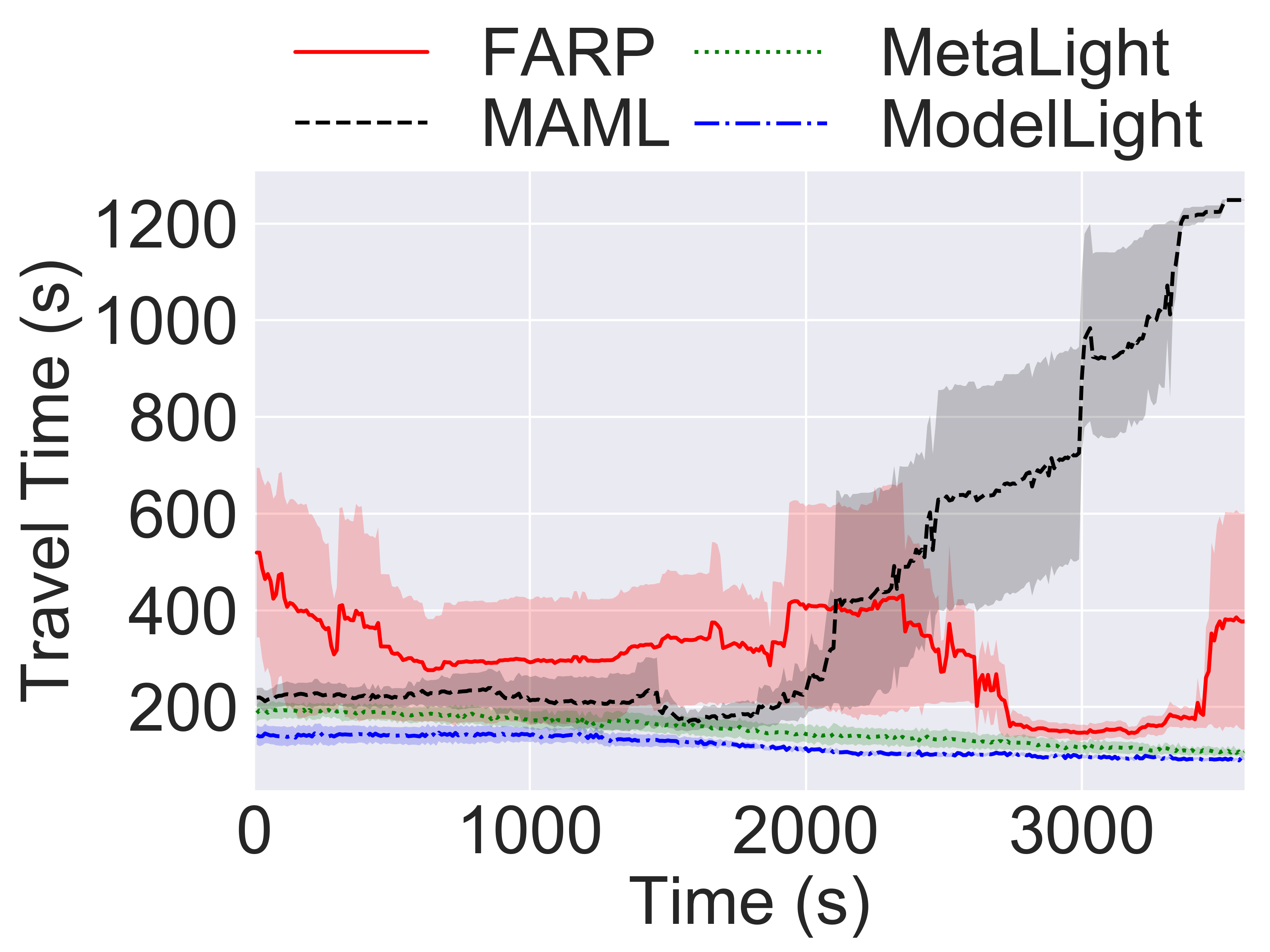}}
  \caption{Comparison of adapted performance on Task 2 and 3. Performance of four scenarios is selected to present. Parts of curves that are out of range are excluded in the selected figures.}
  \label{fig 4} 
\end{figure}

Figures \ref{fig 3} and \ref{fig 4} illustrate learning curves in the meta testing phase for 3 tasks. As we can see, ModelLight shows not only faster convergence rates but also the lowest average travel time among all the methods. Additionally, the variances of performance with 5 random seeds for ModelLight and Metalight are much smaller than those for MAML and FRAP++, which indicates that MAML and FRAP++ are more easily influenced by random seed values while ModelLight and MetaLight are more stable.

\begin{table*}
\scriptsize
\centering
\caption{Adapted performance of different methods on Task 1 (PS1). Each result (mean    $\pm$    standard deviation) represents the average travel time of 2 scenarios and 5 random seeds. The improvement is calculated by comparing ModelLight with the best baseline.}
\label{T1}
\begin{tabular}{lllllll}
\toprule
Phase Setting & 4a             & 4b              & 6a              & 6b              & 6c              & 8               \\ 
\hline
FRAP++          & 96.88 $\pm$ 14.58                  & 259.56 $\pm$ 61.37                 & 413.96 $\pm$ 262.72                 & 724.24 $\pm$ 229.17                 & 294.03   $\pm$ 158.16               & 99.59  $\pm$ 15.97           \\
MAML          & 98.51  $\pm$ 12.43         & 257.38   $\pm$ 61.96        & 353.40  $\pm$ 262.96         & 528.55  $\pm$ 213.03         & 450.07  $\pm$ 97.58         & 112.95  $\pm$ 24.77         \\
MetaLight     & 88.30  $\pm$  6.46         & 284.28  $\pm$ 73.94         & 119.49   $\pm$ 29.15        & 375.64   $\pm$ 101.10        & \textbf{122.89} $\pm$ 11.65 &   98.07  $\pm$ 9.06            \\
ModelLight    & \textbf{80.23}  $\pm$  3.58 & \textbf{245.45}  $\pm$  16.59 & \textbf{114.35}  $\pm$  31.52 & \textbf{325.49}  $\pm$  50.48 & 128.85  $\pm$  17.07          & \textbf{95.29}  $\pm$  11.78  \\ 
\hline
Improvement   & 9.14\%         & 4.64\%          & 4.30\%          & 13.35\%         & -               & 2.83\%          \\
\bottomrule
\end{tabular}
\end{table*}

\begin{table*}
\scriptsize
\centering
\caption{Adapted performance of different methods on Task 2 (PS1) and 3  (PS2). Each result represents the average travel time of scenarios from the same city. ModelLight shows superior performance on all cases. (LA: Los Angeles, AT: Atlanta, JN: Jinan.)}
\label{T2}
\begin{tabular}{llll|lll}
\toprule
\multirow{2}{*}{City} & \multicolumn{3}{c|}{Task 2  (PS1)}                        & \multicolumn{3}{c}{Task 3 (PS2)}                          \\
                      & LA             & AT              & JN              & LA              & AT              & JN               \\ 
\hline
FRAP++                  & 148.80  $\pm$ 38.24                & 271.45  $\pm$ 16.54                & 371.76   $\pm$ 151.29                & 553.60  $\pm$ 194.96                & 532.31     $\pm$ 61.89             & 357.15  $\pm$ 131.67           \\
MAML                  & 134.74  $\pm$ 38.04        & 524.00  $\pm$ 124.50         & 284.31   $\pm$ 102.23        & 686.71   $\pm$ 112.87        & 640.07   $\pm$ 70.33        & 502.28  $\pm$ 98.64          \\
MetaLight             & 115.55  $\pm$ 33.27        & 248.83  $\pm$ 39.47         & 152.64 $\pm$ 40.50          & 402.55  $\pm$ 116.77         & 455.85   $\pm$ 144.88        & 179.56   $\pm$ 47.83         \\
ModelLight            & \textbf{86.28}  $\pm$  7.05 & \textbf{220.74}  $\pm$  21.41 & \textbf{126.37}  $\pm$  14.10 & \textbf{367.30}  $\pm$  35.93 & \textbf{449.28}  $\pm$  115.77 & \textbf{145.70}  $\pm$  16.61  \\ 
\hline
Improvement           & 25.33\%        & 11.29\%         & 17.21\%         & 8.76\%          & 1.44\%          & 18.86\%          \\
\bottomrule
\end{tabular}
\end{table*}

\subsection{Sample Efficiency}
In this experiment, we test how ModelLight performs using only one-tenth of the interactions with the simulator that other meta-learning methods use. Given that the number of interactions with the simulator is fixed for each training round, we train ModelLight only for 10 training rounds and evaluate its performance on the 3 tasks. The best baseline, MetaLight, is used for comparison, with both 100 and 10-round training.

As illustrated in Table \ref{T3} and \ref{T4}, ModelLight trained with only 10 training rounds achieves comparable results to the best baseline (MetaLight trained with 100 training rounds) on 3 tasks, even better results on task 2, which demonstrates the improvement in sample efficiency of ModelLight. In practice, with only one-tenth of the training rounds, we can learn a better initialization for meta test, which further mitigates the reliance on real transitions thus making our proposed method more suitable for traffic signal control issues (\textbf{Q5}).

To eliminate the possibility of overfitting caused by training iterations, we also report the average travel time of MetaLight with 10-round training. It can be seen that MetaLight-10r generates the worst performance, which indicates that MetaLight performs better with more training rounds within a certain range, thus MetaLight trained with 100 rounds can be regarded as the best baseline.

\begin{table*}
\scriptsize
\centering
\caption{Performance of different methods with different training rounds on Task 1 (PS1). ModelLight trained with 10 rounds generates comparable performance to MetaLight (best baseline) trained with 100 rounds.}
\label{T3}
\begin{tabular}{lllllll}
\toprule
Phase Setting    & 4a                         & 4b                                         & 6a                         & 6b                         & 6c                         & 8                           \\ 
\hline
MetaLight-100r & 88.30  $\pm$  6.46         & 284.28  $\pm$ 73.94         & 119.49   $\pm$ 29.15        & 375.64   $\pm$ 101.10        & \textbf{122.89} $\pm$ 11.65 & \textbf{98.07}  $\pm$ 9.06          \\
MetaLight-10r  & 106.77 $\pm$ 22.00 & 498.51 $\pm$ 203.81                 & 248.83 $\pm$ 78.26 & 542.22 $\pm$ 130.94 & 267.16 $\pm$ 75.84 & 227.62 $\pm$  98.59 \\
ModelLight-10r & \textbf{84.81} $\pm$  6.13            & \textbf{269.44}  $\pm$  95.08& \textbf{118.50} $\pm$   28.87          & \textbf{358.91}  $\pm$  68.81  & 158.77  $\pm$  84.19                   & 101.25 $\pm$   22.94                    \\ 
\hline
Improvement      & 3.95\%                     & 5.22\%                                     & 0.83\%                     & 4.45\%                     & -                          & -                           \\
\bottomrule
\end{tabular}
\end{table*}

\begin{table*}
\scriptsize
\centering
\caption{Performance of different methods with different training rounds on Task 2 (PS1) and 3 (PS2). ModelLight outperforms MetaLight when both are trained with 10 rounds in task 2, and shows comparable performance on task 3.}
\label{T4}
\begin{tabular}{llll|lll}
\toprule
\multirow{2}{*}{City} & \multicolumn{3}{c|}{Task 2 (PS1)}                                          & \multicolumn{3}{c}{Task 3 (PS2)}                            \\
                      & LA             & AT                       & JN                       & LA              & AT               & JN                \\ 
\hline
MetaLight-100r         & 115.55  $\pm$ 33.27        & 248.83  $\pm$ 39.47         & 152.64 $\pm$ 40.50          & 402.55  $\pm$ 116.77         & \textbf{455.85}     $\pm$ 144.88          & \textbf{179.56}    $\pm$ 47.83   \\
MetaLight-10r         & 200.89  $\pm$  113.66       & 369.41  $\pm$   120.01                & 355.07  $\pm$   167.37                & 598.58  $\pm$  192.13        & 782.03  $\pm$  183.73         & 390.15   $\pm$    135.85       \\
ModelLight-10r        & \textbf{97.59}  $\pm$   18.20 & \textbf{207.04}  $\pm$  5.50 & \textbf{138.41}  $\pm$  30.87 & \textbf{376.55}  $\pm$  68.78 & 470.11  $\pm$  110.85         & 229.58   $\pm$   99.98        \\ 
\hline
Improvement           & 15.54\%        & 16.79\%                  & 9.32\%                   & 6.46\%          & -                & -                 \\
\bottomrule
\end{tabular}
\end{table*}

\subsection{Ablation Study}
To further study the design of ModelLight, we conduct several experiments via varying the structure or hyperparameters.

\textbf{Ensemble of model vs. single global model} In this paper, an ensemble of intersection models are learned to mitigate the model bias. To verify its effectiveness, we compare ModelLight with an experiment with only one dynamics model. Both methods are trained for 10 training rounds. Results show that ModelLight with an ensemble of models improves performance by an average of 14.54\%, 19.52\% and 14.08\% on Task 1, 2 and 3, respectively.

\textbf{Short rollouts and random starts} The short rollouts method helps alleviate compounding errors and random starts ensure exploration. To examine whether these techniques are helpful and which length of rollouts (denoted as $L$) works the best, we set $L=36$ as default and change $L$ to 10, 120, 360. To eliminate the impact of the total number of imaginary transitions, we keep it the same among these four experiments. 
Therefore, when $L$ is set to 360, the short rollouts method is equivalently abandoned. Results show that $L=360$ performs the worst, $L=10$ is slightly worse, while $L=120$ achieves similar results as $L=36$.

\textbf{Approximator of the dynamics model} A few choices of the dynamics model are evaluated in our study, including Gaussian Process, fully-connected neural network, and LSTM. The experimental results show that ModelLight with LSTM outperforms other methods in terms of travel time with fewer training parameters. 

\section{Conclusion and Discussion}
This paper presents ModelLight, a novel model-based meta-reinforcement learning method  for solving traffic signal control problems. 
ModelLight takes advantage of model-based reinforcement learning and meta-learning to improve the adapted performance and sample efficiency. To mitigate model bias, an ensemble of dynamics models are learned to generate imaginary transitions for policy training. We use short rollouts and random starts for generating imaginary transitions to further reduce compounding model error. Our experiments demonstrate that ModelLight can improve data efficiency compared with other reinforcement learning-based control methods. Furthermore, ModelLight can achieve even better performance with much less interaction with the real environment, which can be very useful in real-world applications. It is also worth noting that the proposed method can be adapted to different intersection types as long as the state and action settings are adjusted accordingly. In the future, we plan to combine transfer learning with ModelLight to further improve the data efficiency and study the performance of ModelLight on more types of intersection models.



\bibliographystyle{IEEEtran}

\begin{thebibliography}{10}
\providecommand{\url}[1]{#1}
\csname url@samestyle\endcsname
\providecommand{\newblock}{\relax}
\providecommand{\bibinfo}[2]{#2}
\providecommand{\BIBentrySTDinterwordspacing}{\spaceskip=0pt\relax}
\providecommand{\BIBentryALTinterwordstretchfactor}{4}
\providecommand{\BIBentryALTinterwordspacing}{\spaceskip=\fontdimen2\font plus
\BIBentryALTinterwordstretchfactor\fontdimen3\font minus
  \fontdimen4\font\relax}
\providecommand{\BIBforeignlanguage}[2]{{%
\expandafter\ifx\csname l@#1\endcsname\relax
\typeout{** WARNING: IEEEtran.bst: No hyphenation pattern has been}%
\typeout{** loaded for the language `#1'. Using the pattern for}%
\typeout{** the default language instead.}%
\else
\language=\csname l@#1\endcsname
\fi
#2}}
\providecommand{\BIBdecl}{\relax}
\BIBdecl

\bibitem{wei2018intellilight}
H.~Wei, G.~Zheng, H.~Yao, and Z.~Li, ``Intellilight: A reinforcement learning
  approach for intelligent traffic light control,'' in \emph{Proceedings of the
  24th ACM SIGKDD International Conference on Knowledge Discovery \& Data
  Mining}, 2018, pp. 2496--2505.

\bibitem{Qadri:2020}
S.~Quadri, M.~Gokce, and E.~Oner, ``State-of-art review of traffic signal
  control methods: challenged and opportunities,'' \emph{European Transport
  Research Review}, vol.~12, p.~55, 2020.

\bibitem{gartner1995development}
N.~H. Gartner, C.~Stamatiadis, and P.~J. Tarnoff, ``Development of advanced
  traffic signal control strategies for intelligent transportation systems:
  Multilevel design,'' in \emph{Transportation Research Record}.\hskip 1em plus
  0.5em minus 0.4em\relax Transportation Research Board, 1995, pp. 98--105.

\bibitem{lowrie1990scats}
P.~Lowrie, ``{SCATS--A Traffic Responsive Method of Controlling Urban Traffic.
  Roads and Traffic Authority, Sydney},'' \emph{New South Wales, Australia},
  1990.

\bibitem{DBLP:conf/nips/OroojlooyNHS20}
\BIBentryALTinterwordspacing
A.~Oroojlooy, M.~Nazari, D.~Hajinezhad, and J.~Silva, ``Attendlight: Universal
  attention-based reinforcement learning model for traffic signal control,'' in
  \emph{Advances in Neural Information Processing Systems, 2020}, 2020.
  [Online]. Available:
  \url{https://proceedings.neurips.cc/paper/2020/hash/29e48b79ae6fc68e9b6480b677453586-Abstract.html}
\BIBentrySTDinterwordspacing

\bibitem{zheng2019learning}
G.~Zheng, Y.~Xiong, X.~Zang, J.~Feng, H.~Wei, H.~Zhang, Y.~Li, K.~Xu, and
  Z.~Li, ``Learning phase competition for traffic signal control,'' in
  \emph{Proceedings of the 28th ACM International Conference on Information and
  Knowledge Management}, 2019, pp. 1963--1972.

\bibitem{zang2020metalight}
X.~Zang, H.~Yao, G.~Zheng, N.~Xu, K.~Xu, and Z.~Li, ``Metalight: Value-based
  meta-reinforcement learning for traffic signal control,'' in
  \emph{Proceedings of the AAAI Conference on Artificial Intelligence}, New
  York, NY, 2020, pp. 1153--1160.

\bibitem{rizzo2019time}
S.~G. Rizzo, G.~Vantini, and S.~Chawla, ``Time critic policy gradient methods
  for traffic signal control in complex and congested scenarios,'' in
  \emph{Proceedings of the 25th ACM SIGKDD International Conference on
  Knowledge Discovery \& Data Mining}, 2019, pp. 1654--1664.

\bibitem{chen2020toward}
C.~Chen, H.~Wei, N.~Xu, G.~Zheng, M.~Yang, Y.~Xiong, K.~Xu, and Z.~Li, ``Toward
  a thousand lights: Decentralized deep reinforcement learning for large-scale
  traffic signal control,'' in \emph{Proceedings of the AAAI Conference on
  Artificial Intelligence}, New York, NY, 2020, pp. 3414--3421.

\bibitem{yumacar}
Z.~Yu, S.~Liang, L.~Wei, Z.~Jin, J.~Huang, D.~Cai, X.~He, and X.-S. Hua,
  ``Macar: Urban traffic light control via active multi-agent communication and
  action rectification,'' in \emph{IJCAI=PRICAI}, Yokahama, Japan, 2020, pp.
  2491--2497.

\bibitem{wang2020large}
X.~Wang, L.~Ke, Z.~Qiao, and X.~Chai, ``Large-scale traffic signal control
  using a novel multiagent reinforcement learning,'' \emph{IEEE Transactions on
  Cybernetics}, vol.~51, pp. 174--187, 2020.

\bibitem{clavera2018model}
I.~Clavera, J.~Rothfuss, J.~Schulman, Y.~Fujita, T.~Asfour, and P.~Abbeel,
  ``Model-based reinforcement learning via meta-policy optimization,'' in
  \emph{Conference on Robot Learning}.\hskip 1em plus 0.5em minus 0.4em\relax
  PMLR, 2018, pp. 617--629.

\bibitem{chua2018deep}
K.~Chua, R.~Calandra, R.~McAllister, and S.~Levine, ``Deep reinforcement
  learning in a handful of trials using probabilistic dynamics models,'' in
  \emph{Advances in Neural Information Processing Systems}, 2018, pp.
  4754--4765.

\bibitem{kurutach2018model}
T.~Kurutach, I.~Clavera, Y.~Duan, A.~Tamar, and P.~Abbeel, ``Model-ensemble
  trust-region policy optimization,'' \emph{International Conference on
  Learning Representations 2018}, 2018.

\bibitem{lambert2019low}
N.~O. Lambert, D.~S. Drew, J.~Yaconelli, S.~Levine, R.~Calandra, and K.~S.
  Pister, ``Low-level control of a quadrotor with deep model-based
  reinforcement learning,'' \emph{IEEE Robotics and Automation Letters},
  vol.~4, no.~4, pp. 4224--4230, 2019.

\bibitem{nagabandi2018learning}
A.~Nagabandi, I.~Clavera, S.~Liu, R.~S. Fearing, P.~Abbeel, S.~Levine, and
  C.~Finn, ``Learning to adapt in dynamic, real-world environments through
  meta-reinforcement learning,'' \emph{International Conference on Learning
  Representations}, 2019.

\bibitem{kaiser2019model}
L.~Kaiser, M.~Babaeizadeh, P.~Milos, B.~Osinski, R.~H. Campbell, K.~Czechowski,
  D.~Erhan, C.~Finn, P.~Kozakowski, S.~Levine \emph{et~al.}, ``Model-based
  reinforcement learning for atari,'' \emph{International Conference on
  Learning Representations}, 2019.

\bibitem{zhang2019solar}
M.~Zhang, S.~Vikram, L.~Smith, P.~Abbeel, M.~Johnson, and S.~Levine, ``Solar:
  Deep structured representations for model-based reinforcement learning,'' in
  \emph{International Conference on Machine Learning}.\hskip 1em plus 0.5em
  minus 0.4em\relax PMLR, 2019, pp. 7444--7453.

\bibitem{finn2017model}
C.~Finn, P.~Abbeel, and S.~Levine, ``Model-agnostic meta-learning for fast
  adaptation of deep networks,'' in \emph{International Conference on Machine
  Learning}.\hskip 1em plus 0.5em minus 0.4em\relax PMLR, 2017, pp. 1126--1135.

\bibitem{genders2020policy}
W.~Genders and S.~Razavi, ``Policy analysis of adaptive traffic signal control
  using reinforcement learning,'' \emph{Journal of Computing in Civil
  Engineering}, vol.~34, no.~1, p. 04019046, 2020.

\bibitem{webster1958traffic}
F.~V. Webster, ``Traffic signal settings,'' Road Research Laboratory, Dept. of
  Scientific and Industrial Research, London, UK, Tech. Rep., 1958, road
  Research Technical Paper No.\ 39.

\bibitem{thorpe1996tra}
T.~L. Thorpe and C.~W. Anderson, ``Tra c light control using sarsa with three
  state representations,'' Citeseer, Tech. Rep., 1996.

\bibitem{abdulhai2003reinforcement}
B.~Abdulhai, R.~Pringle, and G.~J. Karakoulas, ``Reinforcement learning for
  true adaptive traffic signal control,'' \emph{Journal of Transportation
  Engineering}, vol. 129, no.~3, pp. 278--285, 2003.

\bibitem{yang2019cooperative}
S.~Yang, B.~Yang, H.-S. Wong, and Z.~Kang, ``Cooperative traffic signal control
  using multi-step return and off-policy asynchronous advantage actor-critic
  graph algorithm,'' \emph{Knowledge-Based Systems}, vol. 183, p. 104855, 2019.

\bibitem{wei2019colight}
H.~Wei, N.~Xu, H.~Zhang, G.~Zheng, X.~Zang, C.~Chen, W.~Zhang, Y.~Zhu, K.~Xu,
  and Z.~Li, ``Colight: Learning network-level cooperation for traffic signal
  control,'' in \emph{Proceedings of the 28th ACM International Conference on
  Information and Knowledge Management}, 2019, pp. 1913--1922.

\bibitem{zhang2020planlight}
H.~Zhang, M.~Kafouros, and Y.~Yu, ``Planlight: Learning to optimize traffic
  signal control with planning and iterative policy improvement,'' \emph{IEEE
  Access}, vol.~8, pp. 219\,244--219\,255, 2020.

\bibitem{zhu2021meta}
L.~Zhu, P.~Peng, Z.~Lu, X.~Wang, and Y.~Tian, ``Meta variationally intrinsic
  motivated reinforcement learning for decentralized traffic signal control,''
  \emph{arXiv e-prints}, pp. arXiv--2101, 2021.

\bibitem{zhang2020generalight}
H.~Zhang, C.~Liu, W.~Zhang, G.~Zheng, and Y.~Yu, ``Generalight: Improving
  environment generalization of traffic signal control via meta reinforcement
  learning,'' in \emph{Proceedings of the 29th ACM International Conference on
  Information \& Knowledge Management}, 2020, pp. 1783--1792.

\bibitem{langlois2019benchmarking}
E.~Langlois, S.~Zhang, G.~Zhang, P.~Abbeel, and J.~Ba, ``Benchmarking
  model-based reinforcement learning,'' \emph{arXiv preprint arXiv:1907.02057},
  2019.

\bibitem{luo2018algorithmic}
Y.~Luo, H.~Xu, Y.~Li, Y.~Tian, T.~Darrell, and T.~Ma, ``Algorithmic framework
  for model-based deep reinforcement learning with theoretical guarantees,''
  \emph{International Conference on Learning Representations}, 2019.

\bibitem{sutton1990integrated}
R.~S. Sutton, ``Integrated architectures for learning, planning, and reacting
  based on approximating dynamic programming,'' in \emph{Machine learning
  proceedings 1990}.\hskip 1em plus 0.5em minus 0.4em\relax Elsevier, 1990, pp.
  216--224.

\bibitem{rao2009survey}
A.~V. Rao, ``A survey of numerical methods for optimal control,''
  \emph{Advances in the Astronautical Sciences}, vol. 135, no.~1, pp. 497--528,
  2009.

\bibitem{nagabandi2018neural}
A.~Nagabandi, G.~Kahn, R.~S. Fearing, and S.~Levine, ``Neural network dynamics
  for model-based deep reinforcement learning with model-free fine-tuning,'' in
  \emph{2018 IEEE International Conference on Robotics and Automation
  (ICRA)}.\hskip 1em plus 0.5em minus 0.4em\relax IEEE, 2018, pp. 7559--7566.

\bibitem{deisenroth2011pilco}
M.~Deisenroth and C.~E. Rasmussen, ``Pilco: A model-based and data-efficient
  approach to policy search,'' in \emph{Proceedings of the 28th International
  Conference on machine learning (ICML-11)}, 2011, pp. 465--472.

\bibitem{chebotar2017path}
Y.~Chebotar, M.~Kalakrishnan, A.~Yahya, A.~Li, S.~Schaal, and S.~Levine, ``Path
  integral guided policy search,'' in \emph{2017 IEEE international conference
  on robotics and automation (ICRA)}.\hskip 1em plus 0.5em minus 0.4em\relax
  IEEE, 2017, pp. 3381--3388.

\bibitem{tassa2012synthesis}
Y.~Tassa, T.~Erez, and E.~Todorov, ``Synthesis and stabilization of complex
  behaviors through online trajectory optimization,'' in \emph{2012 IEEE/RSJ
  International Conference on Intelligent Robots and Systems}.\hskip 1em plus
  0.5em minus 0.4em\relax IEEE, 2012, pp. 4906--4913.

\bibitem{huang2019learning}
Z.~Huang, W.~Heng, and S.~Zhou, ``Learning to paint with model-based deep
  reinforcement learning,'' in \emph{Proceedings of the IEEE International
  Conference on Computer Vision}, 2019, pp. 8709--8718.

\bibitem{zheng2019diagnosing}
G.~Zheng, X.~Zang, N.~Xu, H.~Wei, Z.~Yu, V.~Gayah, K.~Xu, and Z.~Li,
  ``Diagnosing reinforcement learning for traffic signal control,'' \emph{arXiv
  preprint arXiv:1905.04716}, 2019.

\bibitem{abbeel2006using}
P.~Abbeel, M.~Quigley, and A.~Y. Ng, ``Using inaccurate models in reinforcement
  learning,'' in \emph{Proceedings of the 23rd international conference on
  Machine learning}, 2006, pp. 1--8.

\bibitem{mnih2015human}
V.~Mnih, K.~Kavukcuoglu, D.~Silver, A.~A. Rusu, J.~Veness, M.~G. Bellemare,
  A.~Graves, M.~Riedmiller, A.~K. Fidjeland, G.~Ostrovski \emph{et~al.},
  ``Human-level control through deep reinforcement learning,'' \emph{nature},
  vol. 518, no. 7540, pp. 529--533, 2015.

\end{thebibliography}

\end{document}